# Impact of Data Quality on Deep Neural Network Training

*An Experimental Study*

Subrata Goswami
*sgoswami@umich.edu*

**Abstract**

It is well known that data is critical for training neural networks. Lot have been written about quantities of data required to train networks well. However, there is not much publications on how data quality effects convergence of such networks. There is dearth of information on what is considered good data ( for the task ). This empirical experimental study explores some impacts of data quality. Specific results are shown in the paper how simple changes can have impact on Mean Average Precision (mAP).

## Introduction

Training is arguably the most critical part of building a neural network. The essential inputs to training are the network of course, the parameters and hyper-parameters, and labelled data. Much have been said about quantity of data. However, little is available in terms of quality of data. This paper goes through a number of experiments that artificially introduces errors in the training and test data. This paper focuses on object detection networks. The rationale for using object detection versus say classification, segmentation, etc is a balance between flexibility and ease of manipulation. Changing the characteristics of a bounding box requires significantly less effort than changing shape and pixel properties of a segmentation mask. On the other hand, able to change only the class type of classification does not offer much flexibility.

## Related Works

Over the last few years a number of data sets have developed. Some of the most well known ones are ImageNet, PASCAL-VOC, COCO, KITTI, etc [1,2,3,4]. Most of these datasets have several types of labels - classification, detection, semantic and instance segmentation, multi-modal (e.g. LIDAR and camera) annotations etc. All of these datasets have also been vetted by many for their quality and accuracy. The following table summarizes the accuracy achieved for some of the datasets in object detection.

**Table 1:** Highest Object Detection accuracy on well known data sets [5, 6, 7] .

| Dataset | Accuracy |
|---|---|
| PASCAL-VOC | 92.9(mAP) |
| COCO | 72.9 (IOU 0.5 C17) |
| KITTI | 82.5 ( pedestrian) |

The current highest accuracy on PASCAL-VOC is achieved by an ensemble method [5]. The base accuracy obtained by the Faster RCNN team back in 2015 is 75,5%..

## Experiments

To run the experiments, a data set has to be created or chosen. To validate a new dataset is an expensive process. Hence the choice to use one of the well known object detection data sets. Most open source object detectors use COCO for training. That leaves KITTI or PASCA-VOC, this paper uses PASCAL-VOC.

An object detection network also has to be chosen or developed. Developing a new network from ground up is a very expensive task. Hence the decision to use a fairly well regarded open source network called Tensorpack [7]. The Faster RCNN model pre-trained on COCO from Tensorpack was chosen.

The following are the types of experiments conducted on the PASCAL VOC dataset. All results except for experiment 1, are on the animal subset of images and annotations.

1. All test and validation images and annotations of VOC2012 training and validation.
2. All test and validation animal images and annotations of VOC2012
3. All animal images and annotations with only one object in every annotation file. In this case same image may be used multiple times. The average number of classes per image was observed to be between 1 and 2.
4. Animal images trained on 150 objects per class, and evaluated on 150 objects per class. Also validated on all object annotations set.
5. Same as 4 above, but only 50 objects per class.
6. DIsplacing all the bounding boxes by a fixed number of pixels in both x and y simultaneously by 10
7. Same as 6 but with 20 pixel displacement.
8. Mis-labeling 10% of annotations for each class.
9. Training on all non-difficult as in 2 and testing on: A) non-difficult, non-truncated and frontal pose. B) non-difficult and frontal pose. C) non-difficult and pose left. D) non-difficult and pose right. E) non-difficult and pose unspecified.
10. Training and validation on non-difficult and pose frontal.

## Results

The results of the experiments are detailed in the following.

1. The mAP achieved for all 20 classes of PASCAL was 77.9 after 13 epochs. The best AP was for cat at 90.4 and worst was for diningtable at 57.5 .
2. There are 6 animal classes in the PASCAL VOC data set. The mAP after 13 epochs was 87.1. The highest was for cat at 92.3 and lowest was for cow at 82.4.
3. A) With one object per annotation for both train and test , the mAP achieved was 53.7 after 31 epochs. The best was for cat at

84.9 and wort was for sheep at 23.0. At epoch 13 the mAP was 50.3. B) When the test was done against the more than one object per file annotation ( e.g. test annotations of experiment 2 above), then mAP increases 83.3 at 31 epoch. The highest was for cat at 91.1 and lowest was for sheep at 77.8 .

4. When training and validation were limited to 150 objects, the mAP was 86.6 after 13 epoch. The highest was for cat at 93.8 and lowest for sheep at 77.2.

5. When training and validation were limited to 50 objects, the mAP was 83.0 after 13 epoch. The highest was for horse at 91.9 and lowest for dog at 73.

6. A) When bounding boxes were displaced for both training and validation by 10 pixel in both x and y, the mAP was 82.7. The best was for cat at 91.3 and worst for sheep at 75.0. B) When validation was done against undisplaced annotations (e.g. the test annotation from 2 above), the mAP was 74.8. The best was for cat at 91.2, and worst was for sheep at 56.6.

7. A) When bounding boxes were displaced for both training and validation by 20 pixel in both x and y, the mAP was 68,1. The best was for cat at 84.5 and worst for sheep at 48.0. B) When validation was done against undisplaced annotations ( e.g. the test set from 2 above), the mAP was 56.8. The best was for cat at 84.2, and worst for sheep at 29.1.

8. A) When bounding boxes are mis-labelled by 10 % (e.g. a class other than the correct one, cat for dog) for both train and test data, the mAP was 44.2. The best was for cat at 69.8 and worst for sheep at 18.7. B) When validation was done against the correctly lalled test set, the mAP was 80.9 , with best for horse at 84.4 and worst for cow at 76.3.

9. An experiment with only the non-difficult, non-truncated and frontal pose subset of test images shows an mAP of 61.6 after 13 epochs, with the best for cat at 83.5 and worst for sheep at 40.1. A follow on experiment with non-difficult and frontal pose test images shows an mAP of 63.7, with best for cat at 86.1 and worst for bird at 43,6. Another follow on experiment with non-difficult and pose left showed an mAP of 72.4 , with best for cat at 83.8 and worst for sheep at 58.5 . With non-difficult and pose right the mAP was 67.7 , with best for cat at 87.7 and worst for cow at 53.1. With non-difficult and pose unspecified the mAP was 62.1 , with best for cat at 88.1 and worst for sheep at 32.5.

10. With training and testing on non-difficult and pose frontal, the mAP was 79.6, best for dog at 93.3 and worst for bird at 61.8.

The following table summarizes the above experimental results.

**Table 2: Summary of experiment results.**

| exp | mAP | Best Class | Worst Class |
|---|---|---|---|
| 1 | 77.9 | 90.4 | 57.5 |
| 2 | 87.1 | 92.3 | 82.4 |
| 3A | 53.7 | 84.9 | 23 |
| 3B | 83.3 | 91.1 | 77.8 |
| 4 | 86.6 | 93.8 | 77.2 |
| 5 | 83 | 91.9 | 73 |
| 6A | 82.7 | 91.3 | 75 |
| 6B | 74.8 | 91.2 | 56.6 |
| 7A | 68.1 | 84.5 | 48 |
| 7B | 56.8 | 84.2 | 29.1 |
| 8A | 44.2 | 69.8 | 18.7 |
| 8B | 80.9 | 84.4 | 76.3 |
| 9A | 61.6 | 83.5 | 40.1 |

| 9B | 63.7 | 86.1 | 43.6 |
| 9C | 72.4 | 83.6 | 58.5 |
| 9D | 67.7 | 87.7 | 53.1 |
| 9E | 62.1 | 88.1 | 32.5 |
| 10 | 79.6 | 93.3 | 61.8 |

## Discussion

The results do show some effect on accuracy from data quality. The prominent change observed in experiment 3 is easily explained by how Average Precision (AP) is calculated. There is a penalty for false positives. Hence if the validation annotations only have one object per file, but prediction correctly detects more than one object, then some objects are considered as false positives.

Another prominent change was observed in experiment 8. The reason for the dramatic change here also can be attributed to he how mAP calculation penalizes false positives as in experiment 3.

The mAP did go down a little bit when the number of objects were reduced to 150 (experiment 4). At 50 objects per class (experiment 5), the mAP went down further than it did for 150 objects. The complete object count for training set for each class of animals are as shown in Table 2.

**Table 3:** Distribution of Animal class objects in the PASCAL VOC training set

| class | Object count |
|---|---|
| dog | 756 |
| cat | 605 |
| bird | 560 |
| sheep | 400 |
| horse | 350 |
| cow | 290 |

Displacement of bounding boxes for both training and validation shows decrease in mAP, which accelerates with size of displacement (experiments 6 and 7). This can have significant implications for object detections, as most object detection algorithms uses some form of windowing. Hence, if the follow on regressors are not able to adequately move and shape the bounding boxes, that will result in lower mAP. Interestingly, the accuracy was higher for displaced annotation validation sets (i.e 6A and 6B) compared to the non-displaced. Which implies the network is probably picking up features in the bounding box background that does not belong to the object itself. This issue can be avoided by using instance segmentation instead of object detection, ass background pixels are not included in the mask.

A question needs to be asked about what is preventing close to 100% mAP for the PASCAL-VOC ( and other ) data sets ? To explore that question, first looked at the distribution of objects in the dataset, shown in the following table for the animal classes.

**Table 4: Distribution of animal class objects. Poses are : unspecified, forntal, rear, left, right**

| class | count | difficult | truncated | pose |
|---|---|---|---|---|
| dog | 1598 | 27 | 637 | 640, 452, 48, 226, 232 |
| cat | 1277 | 11 | 621 | 649, 300, 32, 149, 147 |
| bird | 1272 | 102 | 267 | 358, 177, 82, 297, 357] |
| sheep | 1084 | 206 | 371 | 351, 161, 79, 219, 274 |
| horse | 803 | 43 | 317 | 180, 166, 32, 207, 218 |

Experiments 9 A-E were an attempt to see if making the training and validation images more uniform would result is higher mAP. The decrease in mAP in experiment 9A-E looks counter intuitive. This is likely again due to penalty for false positives, where the network detected objects of a class not present in the validation set, as the network was trained on all images.

Experiment 10 trains and validates only on a particular pose - front. The mAP fell short of the mAP beyond the level of Experiment 2. However an improvement was observed for the best AP class, and significant worsening was observed for worst AP class. This could potentially be due to different classes benefitting from different pose.

## Conclusion

A number of tests were done by artificially introducing quality issues. Detrimental impacts were observed in a number of cases. Many more experiments can be done by introducing other types of artificial quality issues. However, even with these limited set of tests, it is evident that for good results, annotations should be precise and complete so that every and all classes of objects in an image are identified. Validation is specially sensitive to lack of precision and accuracy in the validation data set annotation.

## References

1. J. Deng, W. Dong, R. Socher, L.-J. Li, K. Li, and L. Fei-Fei. ImageNet: A Large-Scale Hierarchical Image Database, CVPR09, 2009.
2. Everingham, M., Eslami, S. M. A., Van Gool, L., Williams, C. K. I., Winn, J. and Zisserman, A. , The PASCAL Visual Object Classes Challenge: A Retrospective,International Journal of Computer Vision, 111(1), 98-136, 2015 .
3. Lin TY, Maire M, Belongie S, Hays J, Perona P, Ramanan D, Doll´ar P, Zitnick CL (2014) Microsoft coco: Common objects in context. In: Proc. of the European Conf. on Computer Vision (ECCV)
4. A. Geiger, P. Lenz, C. Stiller, and R. Urtasun. Vision meets robotics: The kitti dataset. International Journal of Robotics Research (IJRR), 2013
5. PASCAL-VOC object detection leaderboard http://host.robots.ox.ac.uk:8080/leaderboard/displaylb.php?challengeid=11&compid=4
6. KITTI object detection leaderboard , http://www.cvlibs.net/datasets/kitti/eval_object.php
7. COCO object detection leader board http://cocodataset.org/#detection-eval
8. Tensorpack, https://github.com/tensorpack/tensorpack